\documentclass{article}
\pdfoutput=1
\usepackage[preprint, nonatbib]{nips_2018}

\usepackage[utf8]{inputenc} % allow utf-8 input
\usepackage[T1]{fontenc}    % use 8-bit T1 fonts
\usepackage{hyperref}       % hyperlinks
\usepackage{url}            % simple URL typesetting
\usepackage{booktabs}       % professional-quality tables
\usepackage{amsfonts}       % blackboard math symbols
\usepackage{nicefrac}       % compact symbols for 1/2, etc.
\usepackage{microtype}      % microtypography
\usepackage[autostyle]{csquotes}

\usepackage{graphicx}
\usepackage{multirow}
\usepackage{subfigure}
\usepackage{verbatim}
\usepackage{amsmath}
\usepackage{enumerate}
\usepackage{color}
\newcommand\norm[1]{\left\lVert#1\right\rVert}

\title{Deep Ranking with Adaptive Margin Triplet Loss}

\author{
  Mai Lan Ha \\
  University of Siegen\\
  \texttt{hamailan@informatik.uni-siegen.de} \\
  \And
  Volker Blanz \\
  University of Siegen \\
  \texttt{blanz@informatik.uni-siegen.de} \\
}

\begin{document}
% \nipsfinalcopy is no longer used

\maketitle

\begin{abstract}
We propose a simple modification from a fixed margin triplet loss to an adaptive margin triplet loss. While the original triplet loss is used widely in classification problems such as face recognition, face re-identification and fine-grained similarity, our proposed loss is well suited for rating datasets in which the ratings are continuous values. In contrast to original triplet loss where we have to sample data carefully, in out method, we can generate triplets using the whole dataset, and the optimization can still converge without frequently running into a model collapsing issue. The adaptive margins only need to be computed once before the training, which is much less expensive than generating triplets after every epoch as in the fixed margin case. Besides substantially improved training stability (the proposed model never collapsed in our experiments compared to a couple of times that the training collapsed on existing triplet loss), we achieved slightly better performance than the original triplet loss on various rating datasets and network architectures.
\end{abstract}

\section{Introduction}

Human perception of particular aspects of images or other stimuli are often measured in terms of comparisons or ratings. For instance, in order to investigate image aesthetics, participants are asked to rate the beauty of images, based on their subjective judgements and criteria. Another example is comparing how much an image is similar to the other image in terms of colors or content categories. From the crowd opinions, we can use machine learning techniques to develop algorithms that reflect this subjective perceptual evaluations. One popular expression for these types of perceptual evaluations is ranking using good similarity or distance metrics. In this work, we study triplet losses used in triplet networks to learn the relationships among images. The shared-weight convolutional network from the triplet network is then used to build a similarity metric for image ranking.

The idea of comparing pairs of images and map their distances from the image space to the feature space was notably introduced in \cite{Chopra_Siamese_2005,Bromley_siamese_1993,Hadsell_siamese_2016}. The similarity metric is built by using two branches of a shared-weight Convolutional Neural Network (CNN) called Siamese Network. The discriminative loss function, aka contrastive loss, is formulated based on whether a pair of input images belong to the same class. The distance between images is therefore small for pairs of the same class and large for pairs from different classes. An extension of Siamese network is triplet networks which consist of three branches of a shared-weight CNN. A triplet network has three input images: one anchor image, one positive example (which is similar to the anchor image) and one negative example (which is different from the anchor image). In feature space, the positive image should have shorter distance to the anchor image than the negative image by a margin $m$, but not necessary very close to the anchor image as in a Siamese Network. This is optimized through a triplet loss function. Triplet loss functions and triplet networks are used to learn fine-grained image similarity and retrieval \cite{Wang_fine_grained_2014,Sohn_nips_2016}. The triplet architecture makes it possible to rank visual similarity for images within the same class. Triplet networks are also used widely in different domains such as face recognition and re-identification \cite{Schroff_facenet_2015,Zhuang_face_2016,Ding_face_2015}, image ranking \cite{Wang_fine_grained_2014,Hoffer_ranking_2015}, aesthetic rating prediction \cite{Schwarz_WACV_2018}, to name a few. 

Even though triplet architectures are being used in various applications, they bear some limitations \cite{Schroff_facenet_2015,Wu_sampling_2017} in the strategy of sampling data for training. If the training data are not sampled carefully, a triplet network is either not efficient in learning, or it tends to collapse. Closely related to training data, the margin $m$ in the loss function is chosen differently and heuristically for different datasets. In many recent rating datasets such as AVA for aesthetic \cite{AVA}, KonIQ10k for quality \cite{koniq10k}, Scenic-Or-Not \cite{ScenicOrNot} for scene beauty and so on, we have more refined data such as users' ratings, the rating distributions and Mean Opinion Scores (MOS). Setting a fixed margin $m$ ignores the refined data. In order to overcome the limitations of the triplet loss and make use of the rating data more efficiently, we propose an adaptive margin triplet loss in which the margin $m$ is adaptive to each input triplet. While the fixed margin triplet loss is used popularly with classification problems such as face re-identification \cite{Schroff_facenet_2015,Zhuang_face_2016,Ding_face_2015} or fine-grained similarity \cite{Wang_fine_grained_2014}, our proposed adaptive margin triplet loss can make use of rating data from rating datasets and therefore increase the ranking performance on these rating datasets.

The fundamental differences between fixed margin triplet loss and our adaptive margin triplet loss is this: with a fixed margin triplet loss, a network learns an embedding space that preserves the orders of the triplets. In our proposed adaptive margin triplet loss, we want to learn an embedding space that preserves not only the triplet orders but also their relative rating distances based on the ground-truth. In other words, we aim for an embedding space that reflects the rating space (users' rating ground-truth). Another main difference between two approaches is on how triplets are utilized for training. In the existing loss, when the relative distances among the triplet exceed the fixed margin, there is no more learning. However, in adaptive margin loss, triplets are still useful to improve the embedding space as long as their distances in the embedding space have not reached their adaptive margins that correspond to their relative distances in the ground-truth ratings.

In summary, the advantage of our proposed loss is the fact that it breaks the curse of training data sampling, stabilizes the training convergence and results in higher accuracies, while still being simple, efficient and intuitive.

% https://github.com/adambielski/siamese-triplet

%We also propose an option to combine adaptive margin triplet loss with regression loss for higher accuracy. %Provide both ranking and better regression accuracies. 
\section{Triplet loss and its challenges} \label{section:triplet_loss}

\subsection{Triplet networks and triplet loss}

\begin{figure}[h!]
\centering
\subfigure[Triplet network architecture for color similarity]{
    \includegraphics[width=0.60\linewidth]{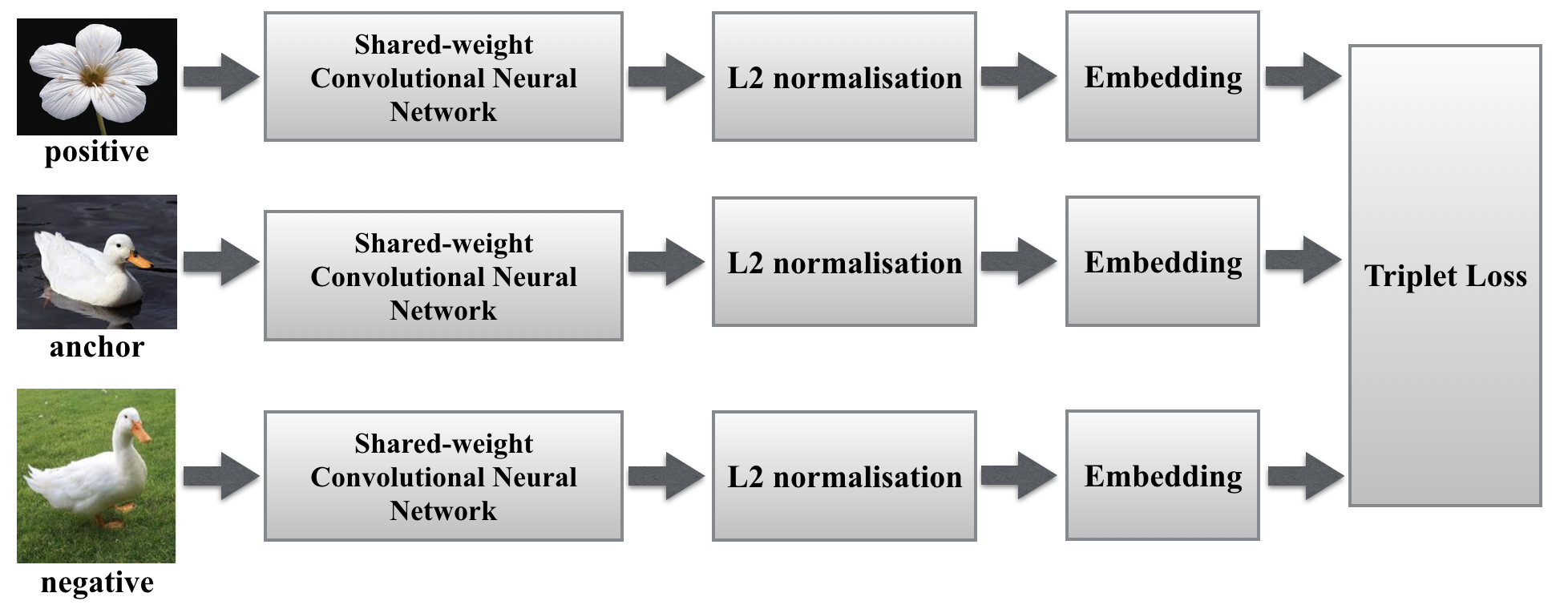}
	\label{fig:triplet_network}
}%
\subfigure[Triplet learning]{
    \includegraphics[width=0.35\linewidth]{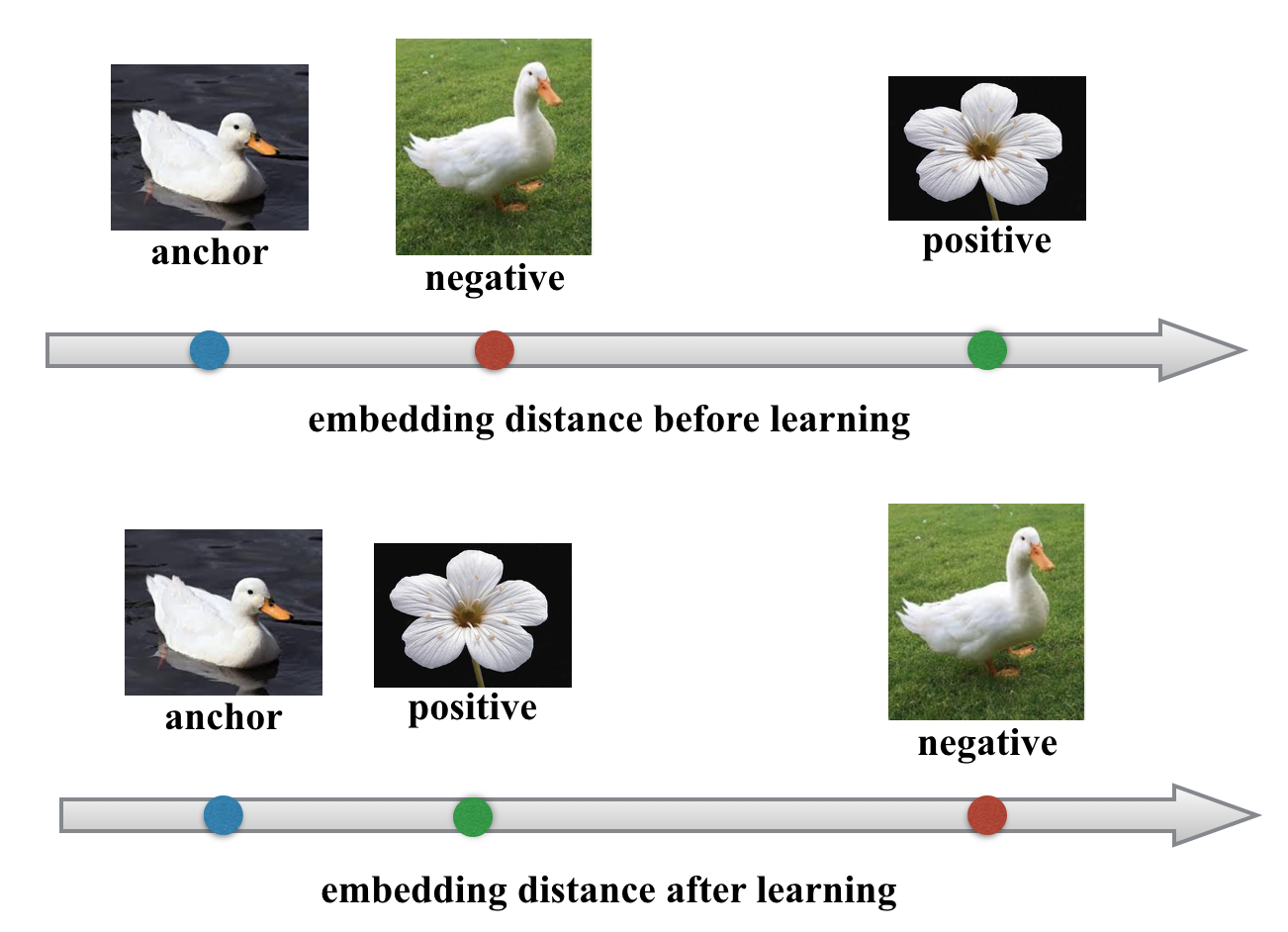}
	\label{fig:triplet_learning}
}
\caption{An overview of a triplet network architecture and triplet learning. In \ref{fig:triplet_network}: it is a triplet network to rank images based on color composition similarity. In terms of color composition, the white flower in black background is more similar to the anchor image than the white duck in green background. In \ref{fig:triplet_learning}, a CNN pretrained on classification is fine-tunned to measure color composition similarity. The embedding distance before training therefore reflects the distance between classes. The distance after retraining or fine-tuning the classification network reflects the color similarity.}
\label{fig:triplet_overview}
\end{figure}

An input image is put through a CNN to be transformed from an image space (such as RGB) to an embedding space. Let $f(x)$ be the embedding of an image $x$. Due to $L2$ normalization \ref{fig:triplet_network}, $\norm{f(x)}_2=1$. Given an anchor image $A$, a positive image $P$ that is similar to the anchor image and a negative image $N$ that is different with the anchor image, the objective is to learn the embeddings of the three input images such that the distance between $A$ and $P$ is smaller than the distance between $A$ and $N$ by a margin $m$:
\begin{equation}
\label{eq:APN_inequality}
	\norm{f(A) - f(P)}_2 + m < \norm{f(A) - f(N)}_2
\end{equation}

The triplet loss for the objective in Eq.~\ref{eq:APN_inequality} is then:
\begin{equation}
\label{eq:triplet_loss}
	L = max(\norm{f(A) - f(P)}_2 - \norm{f(A) - f(N)}_2 + m, 0)
\end{equation}

For the embedding distance, we can use $\norm{.}^2_2$ as in \cite{Schroff_facenet_2015} or simply just $\norm{.}_2$ as suggested in \cite{Wu_sampling_2017} to reduce the severity of network collapsing. In our experiments, we consitently use the latter one. The margin $m$ is fixed at a heuristic value based on what works best on a given dataset.

\subsection{Challenges} \label{subsection:triplet_challenges}

Even though triplet architecture is being used in various applications and currently holds the state-of-the-art performances for embedding distance metrics, it bears some limitations \cite{Schroff_facenet_2015,Wu_sampling_2017,Sohn_nips_2016,Hermans_defense_2017}:

\begin{enumerate}[i]
	\item \textbf{Training data sampling:} If the negative image is very different from the anchor image and positive image such that $\norm{f(A) - f(N)}_2$ is much larger than $\norm{f(A) - f(P)}_2$, the loss $L=0$ (Eq.\ref{eq:triplet_loss}). Thus, no learning happens. Some strategies such as hard negative mining or semi-hard negative mining are proposed. However, mining hard triplets for training is time consuming because the whole set of training data has to be re-evaluated after every epoch to re-compute their embedding distances. Another unfortunate fact is that hard negative mining easily results in model collapses such that all the embeddings are mapped to the same value. 
    
% * <subpic@gmail.com> 2018-05-17T19:19:15.496Z:
% 
% Explain in a few words what mining means, for those that are not familiar with the topic?
% 
% ^.

	\item \textbf{Choice of constant margin $m$:} the choice of the margin $m$ also plays an important role for the learning efficiency and, in some cases, it can also lead to model collapsing. Depending on the properties of the datasets, different values of $m$ should be chosen heuristically.
% * <subpic@gmail.com> 2018-05-17T19:26:49.379Z:
% 
% Imply how hard it is to choose a margin, to show that it's an important point. Some ideas for arguing:
% - how many iterations are needed to find a good one?
% - how do people go about doing it, and how sensitive is the choice of margin?
% - how sparse is the set of good options? etc
% 
% ^.

	\item \textbf{Relative distance:} the distances between the negatives and the anchors are only required to be larger than the distances between positives and the anchors by a fixed margin $m$. However, the additional information in the absolute rating values is not fully exploited in rating datasets. 

\end{enumerate}

\section{Adaptive Margin Triplet Loss} \label{section:margin_loss}

As described in Section \ref{section:triplet_loss}, the margin $m$ in the triplet loss is fixed for all triplets. In this section, we propose to change the fixed margin $m$ to be adaptive based on the training data. We will later analyze the advantages of this adaptive margin.

\begin{figure}[h!]
\centering
\subfigure[Embedding learning with fixed margin $m$]{
    \includegraphics[width=0.5\linewidth]{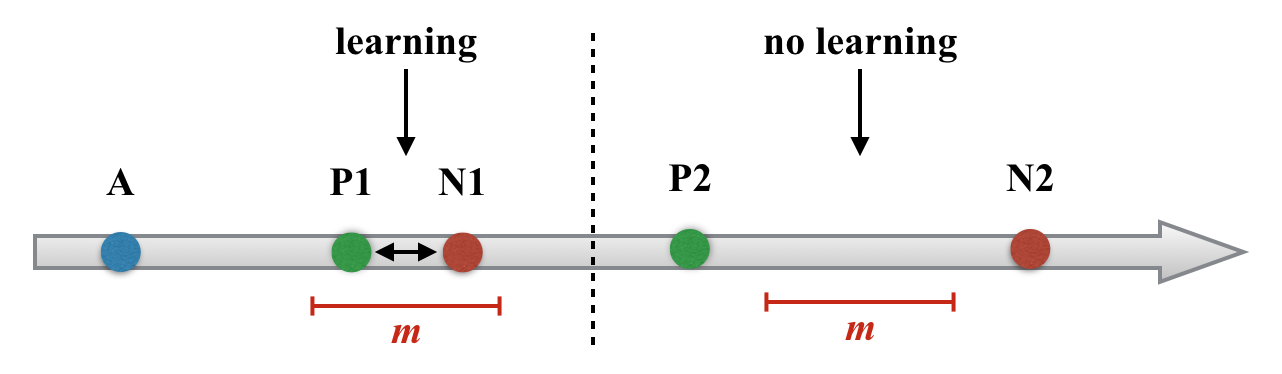}
	\label{fig:fixed_m_learning}
}%
\subfigure[Embedding learning with adaptive margin]{
    \includegraphics[width=0.5\linewidth]{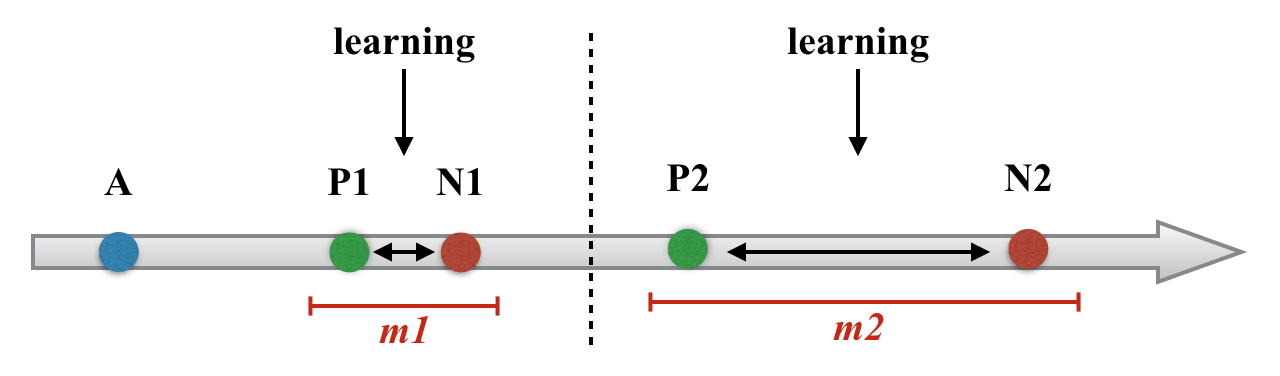}
	\label{fig:adaptive_m_learning}
}
\caption{A comparison on learning embedding with fixed margin and adaptive margin. In the case of fixed margin \ref{fig:fixed_m_learning}, the network will not learn anything if the distance between $P2$ and $N2$ is already larger than the fixed margin $m$. However, with adaptive margin \ref{fig:adaptive_m_learning}, every triplet has its own adaptive margin based on their ratings from the dataset. Therefore, even though the distance between $P2$ and $N2$ is large, the network still has a chance to continue learning, and it is stabilized.}
\end{figure}

\subsection{Adaptive margin on rating datasets}

 Given a triplet data $\{A_i, P_i, N_i\}$. Assume that the rating values are from $1$ to $n$. The adaptive margin for this triplet data is defined as:

 \begin{equation}
 \label{eq:adaptive_margin}
 	m_i = \left|\frac{ d_{GT}(A_i,P_i)- d_{GT}(A_i,N_i)}{n-1}\right| \\
 	%d(A,P) = |S(A_i) - S(P_i)| \\
 	%d(A,N) = |S(A_i) - S(N_i)|
 \end{equation}

where $d_{GT}(A_i,P_i)$ is the ground-truth distance between anchor $A_i$ and positive $P_i$, $d_{GT}(A_i,N_i)$ is the ground-truth distance between anchor $A_i$ and negative $N_i$ in the rating space. By dividing by $n-1$, $m_i$ is normalized to $[0,1]$. In datasets such as AVA \cite{AVA} and KonIQ-10k \cite{koniq10k}, the users rated individual images. Therefore, $d_{GT}(A_i,P_i) = |MOS(A_i) - MOS(P_i)|$ and $d_{GT}(A_i,N_i) = |MOS(A_i) - MOS(N_i)|$ where $MOS(A_i)$, $MOS(P_i)$, $MOS(N_i)$ are the Mean Opinion Scores of $A_i, P_i, N_i$ respectively. In other datasets where the ratings are evaluated by comparing pairs of images, $d_{GT}(A_i,P_i)$ and $d_{GT}(A_i,N_i)$ are MOS values for pairs $(A_i,P_i)$ and $(A_i,N_i)$ respectively.

\subsection{Implementation of adaptive margin}

\begin{figure}[h!]
\centering
\includegraphics[width=0.7\linewidth]{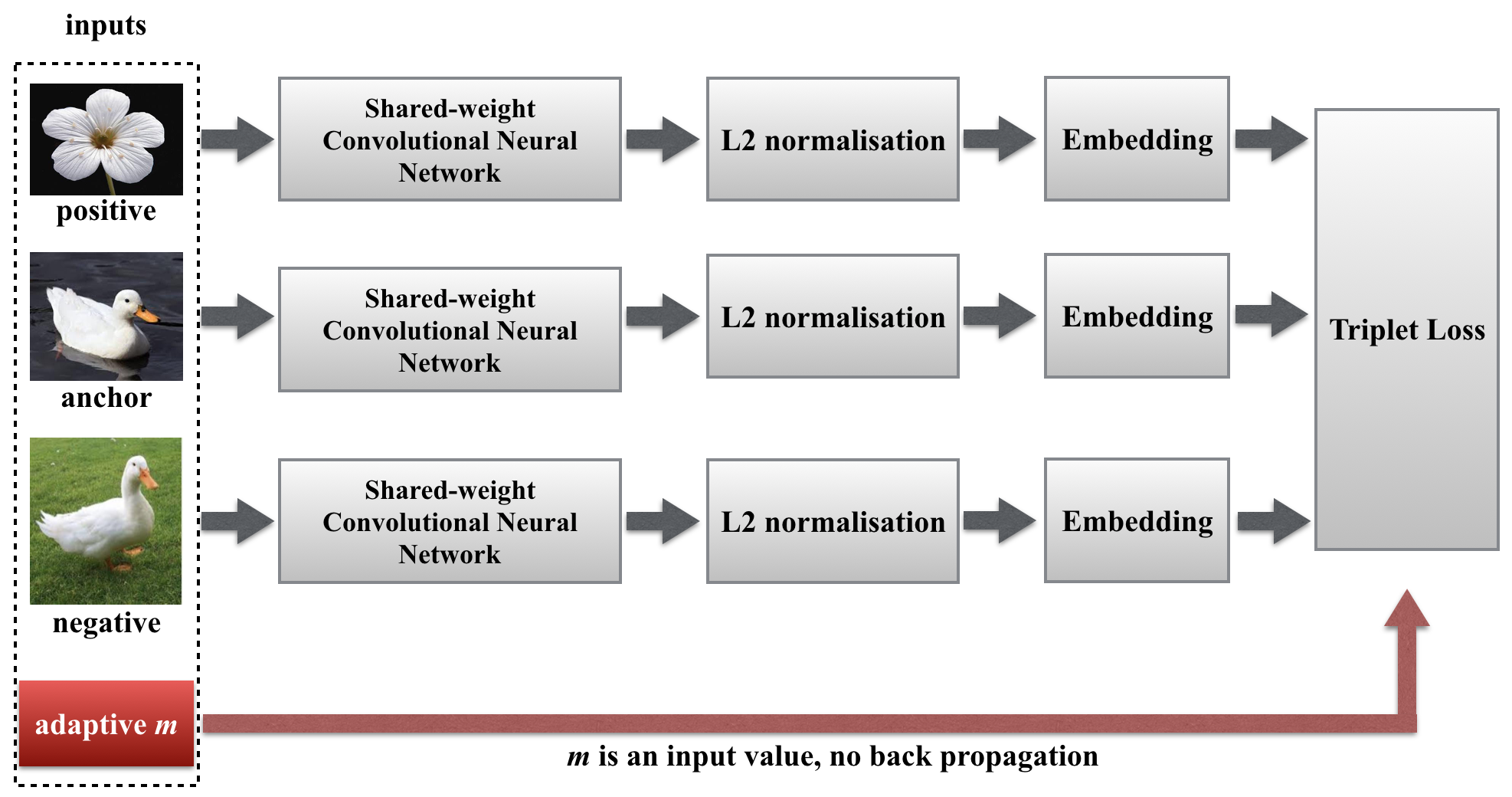}
\caption{Triplet network architecture with adaptive margin for the triplet loss.}
\label{fig:adaptive_m_implementation}
\end{figure}

We implement adaptive margin for triplet loss using keras tensorflow. In the implementation, $m$ is an input variable that changes together with the input triplet data (Fig.~\ref{fig:adaptive_m_implementation}). Each training data is a quadruplet $\{A_i, P_i, N_i, m_i\}$. The value of $m$ for each triplet is computed using Eq.~\ref{eq:adaptive_margin} and then used to compute the loss in Eq.~\ref{eq:triplet_loss}. The backpropagation is on the CNN but not on $m$. 
% * <subpic@gmail.com> 2018-05-17T19:28:34.471Z:
% 
% The figure is a bit confusing to me:
% 
% Is the adaptive margin really part of the input? For instance, a MOS value is a target that is used in the loss (e.g. L2). It looks to me that the adaptive margin is a function of the MOS, thus it should be simply part of the loss and not an input to the network? 
% 
% ^.

\subsection{Advantages of adaptive margin}

Our proposed adaptive margin alleviates the problems with a fixed margin discussed in Section \ref{subsection:triplet_challenges}:

\begin{enumerate}[i]
	\item \textbf{Training data sampling:} Instead of taking care of the hard mining problem, with adaptive margin, we can generate as many quadruplet data $\{A_i, P_i, N_i, m_i\}$ as we want before the training. Therefore, it reduces the time of generating hard (semi-hard) triplets after every training epoch. 

	\item \textbf{Choice of constant margin $m$:} We are free from choosing and fine-tuning $m$.

	\item \textbf{Relative distance:} In rating datasets, the absolute rating values are efficiently utilized. Therefore, we also have more training data.

	\item \textbf{Training stability and efficiency:} From many experiments, it shows that adaptive margin in triplet loss is a great help in avoiding model collapsing. With small datasets that have few ten thousands of images, generating quadruplets $\{A_i, P_i, N_i, m_i\}$ leads to much more training data compared to hard and semi-hard mining. Therefore, the ranking accuracies from adaptive margin are also higher than from fixed margin. The comparison and analysis of numerical results are shown in Section \ref{section:results}.
% * <subpic@gmail.com> 2018-05-17T19:33:49.853Z:
% 
% Maybe it can be interesting to say just how little data there is when you have to abide by the triplet selection rules for a fixed margin. For instance, assuming a distribution of ratings like in KonIQ, how many triplets could you form?How many are needed to train well? (Estimate)
% 
% ^.

\end{enumerate}

\section{Experimental Results} \label{section:results}

We compare the performances of the fixed margin triplet loss and the proposed adaptive margin triplet loss on different datasets and different network architectures. Our goal is not to produce the best performance for each dataset but rather to highlight the differences between the two losses. Therefore, we use simple and various network architectures, from transfer learning, fine-tuning to training from scratch. In order to have fair comparisons, we use exactly the same architectures, parameters and training data for both losses in each evaluation.

The best suited dataset that a triplet network can be used on is a dataset that has precise measurements of distances between anchor and positive images as well as the distances between anchor and negative images. In this work, we introduce a dataset called COLOR-SIM to measure the perceptual color composition similarity between pairs of images. We then compare the performance of fixed margin and adaptive margin losses on this dataset.

\subsection{Perceptual Color Composition Similarity Dataset (COLOR-SIM)} 

COLOR-SIM is a dataset for rating perceptual color composition similarity using crowd-sourcing. Given a pair of images, observers rate the degree to which it is similar with respect to colors. Statistically, numbers of pairs that are totally different in colors are much higher than similar pairs. Therefore, we use an active learning approach in two stages to select images for which ratings for color similarity make sense. 

In the first stage, we create a small set of image pairs for which similarity is clearly defined: either very similar (3519 pairs) or completely dissimilar (3519 pairs) from INRIA Holidays \cite{Jegou2008_Holiday} and Pixabay \cite{Pixabay} datasets. From this binary seed dataset, we train a binary classification network (\textit{seed-net}) to identify similar/dissimilar labels for pairs of images. We then use the \textit{seed-net} to sample more image pairs and have them annotated by users as similar or dissimilar. The new annotated pairs are used to improve \textit{seed-net} performance. In the second stage, we use the \textit{seed-net} to build a dataset for user rating on a finer-grained 5-point scale where 1 means a pair is totally different and 5 means the images are very similar or almost identical. The rating dataset consists of 31.248 pairs of images (1.302 reference images, 24 evaluated images for each reference). It is important to choose evaluated images such that the ratings are presented for all 5 rating options. For each reference, we choose images in the binary seed dataset which contains similar/dissimilar labels for rating options $\{1,5\}$ and selections of top results from the improved \textit{seed-net} for rating options $\{2,3,4\}$.

\subsection{Results on COLOR-SIM dataset}

We split the COLOR-SIM dataset into a training set (24840 pairs) and a test set (6210 pairs). There are no common reference images in the two sets. We use the training set for training and validating, and evaluate the performance by computing SROCC on the test set. SROCC is the Spearman rank order correlation between the predicted results and users' ratings. 

\begin{table}
  \caption{SROCC ranking results on COLOR-SIM dataset}
  \label{table:color-sim_results}
  \centering
  \begin{tabular}{cccc}
    \toprule
    %\multicolumn{2}{c}{Part}                   \\
    %\cmidrule(r){1-2}
    									& MOS regression    & Fixed $m=0.5$  & Adaptive margin (ours)	\\
    \midrule
    Transfer learning using VGG19 fc7 	& 0.780  			& 0.787 			& \textbf{0.846}    			\\
    ConvNet								& 0.866    			& 0.816 			& \textbf{0.872}      		\\
 
    \bottomrule
  \end{tabular}
\end{table}

\textbf{Triplet generation:} Instead of choosing only hard triplets, we use all the rating data in COLOR-SIM dataset. Each reference image is used as an anchor image. All 24 evaluated images that are paired with a reference image are used to create positive and negative examples. We have totally over 280K triplets.

\textbf{Similarity vs. distance:} With the COLOR-SIM dataset, users' tasks are to rate image similarity in terms of color composition. The rating scales are from 1 to 5 where 1 is totally dissimilar and 5 is very similar to identical. Therefore, the distance between a pair of images $(x,y)$ is computed as $d(x,y) = (5 - s(x,y))/4$ where $s(x,y)$ is the similarity score of $(x,y)$. The division by 4 is to normalize $d(x,y)$ to $[0,1]$. Hence, a positive image will have a high similarity score and a low distance to the anchor image.

\textbf{Evaluation:} We compute the distances between pairs of test images in the embedding space. The ranking accuracy is the SROCC between these computed distances and MOS from users' ratings.

In this dataset, we use two training strategies. One is transfer learning using fc7 features from the VGG19 network \cite{simonyan14_vgg19}. The other strategy is to train a CNN (ConvNet) from randomized initializations. ConvNet has the same convolutional kernel parameters as AlexNet \cite{NIPS2012_AlexNet}. Unlike AlexNet, the ConvNet does not have any cross channel normalization or batch normalization for simplicity. We compare the ranking SROCC performances among Mean Opinion Score (MOS) regression, triplet loss with fixed margin and our proposed triplet loss with adaptive margin. The results in Table \ref{table:color-sim_results} show that adaptive margin has the best performance. We also notice that ConvNet gives better ranking results than transfer learning from VGG19. The reason is because content objects and colors are not highly correlated. VGG19 is pre-trained for image classification. The fc7 features are then tailored for good classification performance rather than for visual similarity, which includes colors.

% \begin{enumerate}[i]

% 	\item \textbf{Triplet generation:} Instead of choosing only hard or semi-hard triplets, we use all the rating data of the whole COLOR-SIM dataset. Each reference image is used as an anchor image. All 24 evaluated images of a reference image are used to create positive and negative examples. We have totally over 280K triplets.

% 	\item \textbf{Similarity vs. distance:} With COLOR-SIM dataset, users' tasks are to rate image similarity in terms of color composition. The rating scales are from 1 to 5 where 1 is totally dissimilar and 5 is very similar to identical. Therefore, the distance between a pair of images $(x,y)$ is computed as $d(x,y) = (5 - s(x,y))/4$ where $s(x,y)$ is the similarity score of $(x,y)$. The division by 4 is to normalize $d(x,y)$ to $[0,1]$. Therefore, a positive image will a have high similarity score and a low distance to the anchor image.
    
%     \item \textbf{Evaluation:} We compute the distances between pairs of test images $(x,y)$ in the embedding space. The ranking accuracy is the SROCC correlation between these computed distance and MOS of users' ratings.

% \end{enumerate}

\subsection{Evaluations on single image rating datasets}

% Among many single image rating datasets, we choose KonIQ-10k \cite{koniq10k} and AVA \cite{AVA} datasets for evaluation the performance of triplet losses. KonIQ-10k is a rating dataset for image quality and contains over 10K images. AVA is an aesthetic rating dataset that contains over 250K images.

In single image rating datasets, users rate each image individually according to some criterion. For example, in AVA \cite{AVA}, AADB \cite{kong2016_AADB} and ScienicOrNot \cite{ScenicOrNot} users rate how beautiful each image is.  Among many single image rating datasets, we choose KonIQ-10k \cite{koniq10k} datasets for evaluating the performance of triplet losses. KonIQ-10k is a rating dataset for image quality and contains over 10K images. The reason why we choose this dataset is because the user ratings are studied thoroughly and are also compared to expert ratings \cite{Hosu2018-expertise-screening}. Unlike the COLOR-SIM dataset, where the ratings are comparisons between pairs of images, these dataset ratings are on individual images. Therefore, we have to generate triplets and evaluate the ranking differently.

\subsubsection{KonIQ-10k dataset} \label{section:KonIQ-10k}

\begin{figure}[h!]
\centering
\subfigure[]{
    \includegraphics[width=0.30\linewidth]{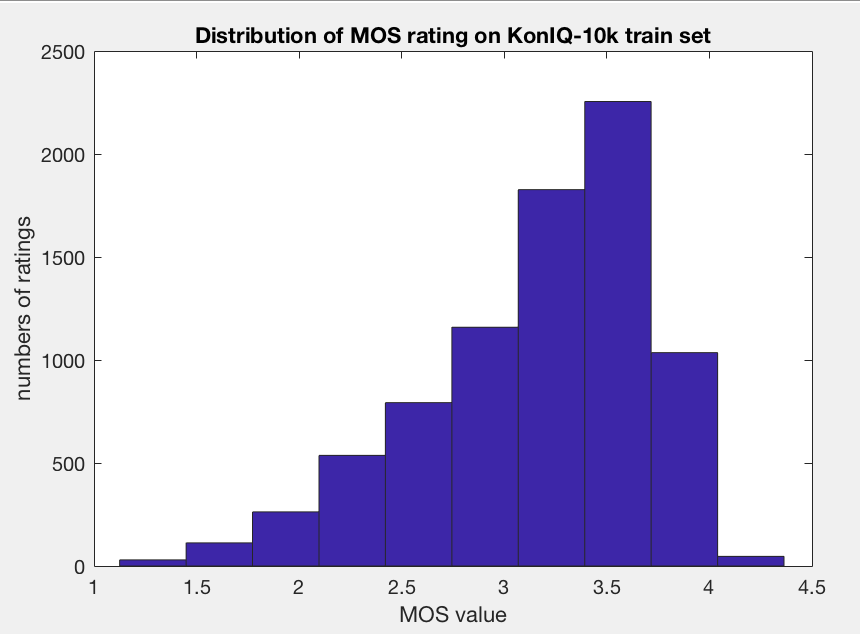}
	\label{fig:koniq10k_MOS_train}
}%
\subfigure[]{
    \includegraphics[width=0.3\linewidth]{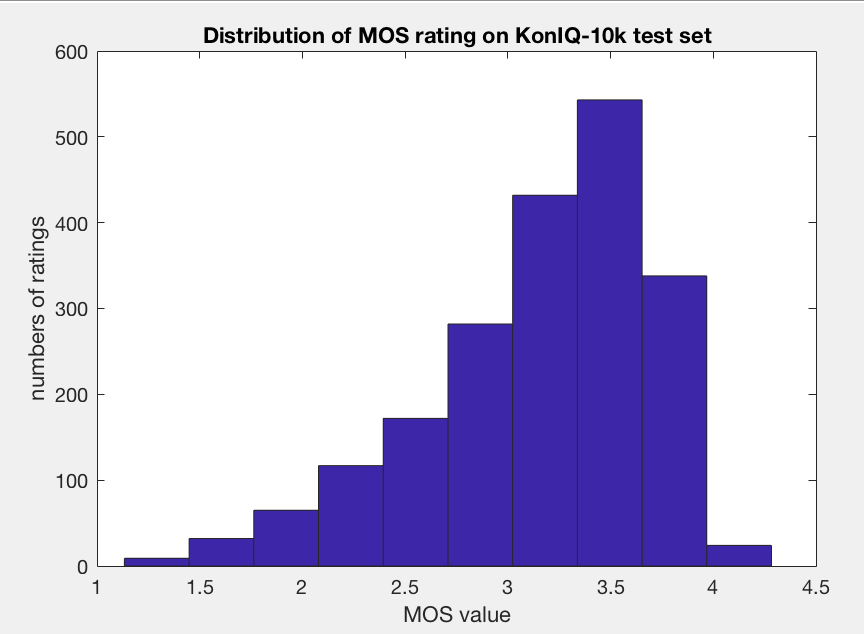}
	\label{fig:koniq10k_MOS_test}
}
\subfigure[]{
    \includegraphics[width=0.3\linewidth]{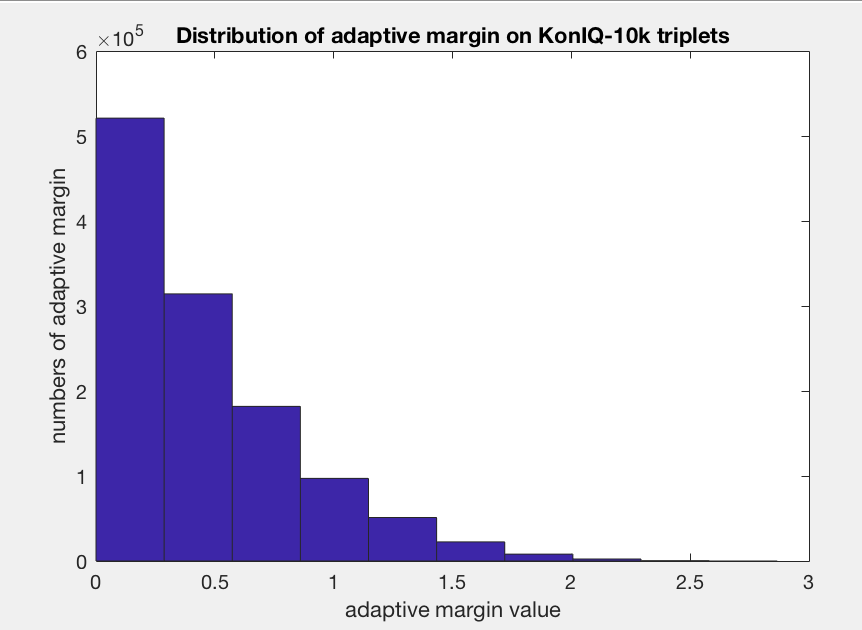}
	\label{fig:koniq10k_margin_triplet}
}
\caption{KonIQ-10k data statistic. \ref{fig:koniq10k_MOS_train}: the distribution of MOS values in the 8K training set. \ref{fig:koniq10k_MOS_test}: the distribution of MOS values in the 2K testing set. The two rating distributions from the training set and testing set are similar. \ref{fig:koniq10k_margin_triplet}: the distribution of adaptive margin on the training triplets. The adaptive margin reflects the relative distances between $d(A,P)$ and $d(A,N)$. Even though the data are randomly sampled, we have enough hard and semi-hard triplet examples.}
\label{fig:koniq10k_distribution}
\end{figure}

\textbf{Triplet generation:} We first split the KonIQ-10k dataset into an 8K training set and a 2K test set. The training triplets are generated from the 8K training set. We select all the images in the 8K training set as anchors for the triplets. With each anchor, we randomly select 300 images in the 8K set to form 150 triplets. As a result, we have more than 1 million triplets for training. The statistics on MOS rating values and adaptive margin are shown in Fig.~\ref{fig:koniq10k_distribution}. The adaptive margin reflects the relative distances between $d(A,P)$ and $d(A,N)$. Fig.~\ref{fig:koniq10k_margin_triplet} shows that even though the data are randomly sampled, we have enough hard and semi-hard triplet examples.

\textbf{Evaluation:} We use $m=0.5$ as a fixed margin and Eq.\ref{eq:adaptive_margin} to compute the adaptive margins. In triplet training, we optimize for the distance between pairs of images in embedding spaces, but not the MOS rating score of each individual image. The ranking evaluation is therefore not straight forward as with the COLOR-SIM dataset of which the MOS ratings are measured relatively between image pairs. There are several options to evaluate the single rating dataset:

\begin{itemize}
	\item Choose one image which has the highest or lowest MOS rating as reference. The ranking of the test set is then relative to this reference image. We choose this method to report the results in Table \ref{table:KonIQ-10k_results}.
	\item Add constraints such that $||f(x)|| = MOS(x)$ where $f(x)$ is the embedding vector of $x$.
	\item Add constraints on the direction between anchor and negative images as suggested in \cite{Schwarz_WACV_2018}.
	\item Add a regression loss on MOS for the anchor image. We explore this option in Section \ref{section:combine_losses}.
\end{itemize}

\begin{table}[h!]
  \caption{SROCC ranking results on KonIQ-10k dataset}
  \label{table:KonIQ-10k_results}
  \centering
  \begin{tabular}{cccc}
    \toprule
    %\multicolumn{2}{c}{Part}                   \\
    %\cmidrule(r){1-2}
    									& MOS regression    & Fixed $m=0.5$  & Adaptive margin (ours)	\\
    \midrule
    Transfer learning using VGG19 fc7	& 0.466  			& 0.645 			& \textbf{0.664}    	\\
    Inception V3 (input size 512x384)	& 0.426				& 0.758				& \textbf{0.768}	\\
    VGG19 conv5 (input size 512x384		& 0.482    			& 0.799 			& \textbf{0.806}      	\\
%     ConvNet (input size 512x384)		& -					& -				& -						\\
 
    \bottomrule
  \end{tabular}
\end{table}

The results in Table \ref{table:KonIQ-10k_results} shows that adaptive margin has the best ranking accuracies accross different networks. KonIQ-10k is a dataset to evaluate image quality. We see that by using VGG19 fc7 features, the input images have to be 224x224. By rescaling an image from size 1024x768 to 224x224, the quality is reduced, so the ranking accuracy is low. However, by using images with input size 512x384, the accuracy is improved significantly. We use InceptionV3 \cite{Szegedy2016_Inception} and VGG19 \cite{simonyan14_vgg19} networks without the top fully-connected layers, with an input image size 512x384 and Global Average Pooling (GAP) to pool the final convolutional feature maps. The results can be further improved by fine-tuning through the whole network. However, our goal is not to provide the best performance networks, but to compare the performances between fixed margin and adaptive margin triplet losses.

Having only 8K images for training, the MOS regression results are consistently low in all the networks. However, using triplets, we can still generate a large amount of training data. By optimizing for the embedding space using triplet losses, we map these 8K images to the embedding hypersphere such that the relative distances among them agree with the MOS rating distances.

\subsubsection{AVA-10k dataset} % \textcolor{red}{to be finished}}

AVA dataset is an aesthetic rating dataset. It contains 250K images. We subsample to 25K, 50K and 80K subsets for training and evaluation. We split 80-20 for training and testing. The comparison results among MOS regression, fixed margin and adaptive margin losses are presented in Table \ref{table:AVA_results}.

\begin{table}[h!]
  \caption{SROCC ranking results on subsets of AVA dataset}
  \label{table:AVA_results}
  \centering
  \begin{tabular}{cccc}
    \toprule
    %\multicolumn{2}{c}{Part}                   \\
    %\cmidrule(r){1-2}
    									& MOS regression    & Fixed $m=0.5$  & Adaptive margin (ours)	\\
    \midrule
    VGG19 conv5 AVA-25k 		& 0.231  			& 0.333 			& \textbf{0.460}    	\\
    VGG19 conv5 AVA-50k			& 0.282				& 0.514				& \textbf{0.520}	\\
 	VGG19 conv5 AVA-80k			& 0.312				& 0.519				& \textbf{0.530}	\\
    \bottomrule
  \end{tabular}
\end{table}

The performances of state-of-the-art methods on aesthetic ranking are in Table \ref{table:AVA_results_compare}. Our best result is $0.530$ compared to $0.558$ from \cite{kong2016_AADB}. First of all, we selected only less than 1/3 of the AVA dataset, and we only do transfer learning without fine-tuning the whole network. We can observe the effect of having less data by comparing the MOS regression results. In AVA-80K, the SROCC of MOS regression is $0.312$ compared to the regression SROCC $0.4995$ on the whole dataset from the paper \cite{kong2016_AADB}. The performance of adaptive margin loss increases when the size of the training data increases.

\begin{table}[h!]
  \caption{SROCC ranking results of state-of-the-art methods.}
  \label{table:AVA_results_compare}
  \centering
  \begin{tabular}{ccc}
    \toprule
    %\multicolumn{2}{c}{Part}                   \\
    %\cmidrule(r){1-2}
    Methods															&  ACC (\%)  &  SROCC $\rho$\\
    \midrule    
    Regression \cite{Schwarz_WACV_2018} 							& -			 & 0.4995	\\
%    Regression (our subset of data)									& -			 & 0.282	\\
    Triplet loss + directional loss \cite{Schwarz_WACV_2018} 				& 75.83\%    & - 		\\
    Reg + Rank + Att + Cont \cite{kong2016_AADB} 			& 77.33\%    &	0.5581	\\
%    Adaptive margin (ours)											& -    		 & 0.520   	\\
 
    \bottomrule
  \end{tabular}
\end{table}

% \subsection{Model collapsing analysis}

% Convergence between fixed margin vs adaptive margin (around fixed margin vs always converge).

% Add figures for:

% 1. Dataset characteristics (plot the histogram on the pair distances of koniq10k and color-sim for comparison)

% 2. Figure to display the collapsing

% 3. Figure to display the none-collapsing when reduce fixed $m$ and same case for adaptive margin

% \subsection{Discussion}

% - Propose an adaptive margin triplet loss that used in triplet network to do ranking on rating datasets.

% - Summary about the performance

% - Some conclusion about model collapsing

% - The performances can be increased for each dataset by fine-tune more futher back to the whole base network architecture or more complicated network architecture. The performance can also be increased by data augmentation.

% - The point is not to find a network architectures that give the best performance in each dataset but to compare the performance of fixed margin vs adaptive margin given that they use the same network architectures, network params and training params.

\section{Combining Ranking and Regression Prediction} \label{section:combine_losses}

\begin{figure}[h!]
\centering
\includegraphics[width=0.98\linewidth]{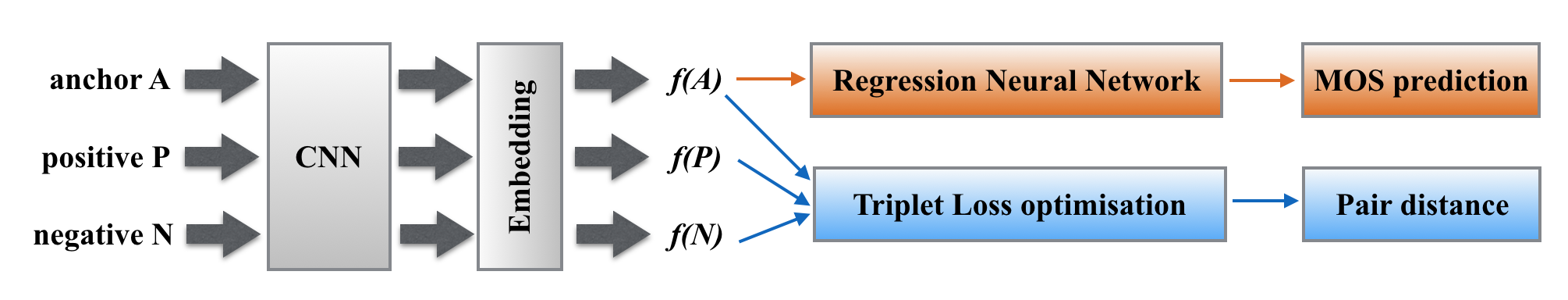}
\caption{Combining triplet loss and regression loss for single image rating datasets.}
\label{fig:combining_losses}
\end{figure}

The disadvantage of triplet ranking networks is that they do not predict rating values. In rating datasets for image-pair comparison such as COLOR-SIM, relative distances in the embedding space reflect the pair ratings. That is why triplet models fit naturally with our COLOR-SIM dataset. In single image rating datasets, we can derive relative distances between pairs of images to generate triplets for embedding training. However, the embedding space is not optimized for the rating values of individual images. A problem occurs when we try to infer rating values for individual test images to evaluate the ranking. Several solutions are described in Section \ref{section:KonIQ-10k}. Here we propose a solution that combines triplet losses with regression loss. The total loss is computed as follows:

\begin{align*}
	L_{total} &= \alpha * L_{triplet} + \beta * L_{regression} 	\\	
	L_{triplet} & = max(\norm{f(A) - f(P)}_2 - \norm{f(A) - f(N)}_2 + m, 0) \\
	L_{regression} &= |g(f(A) - MOS|
	% L_{regression} &= \begin{cases} |d(f(A),f(P)) - MOS|, & \text{for pair rating datasets} \\
									% |g(f(A) - MOS|, & \text{for individual image rating datasets}
						% \end{cases}
\end{align*}

where $\alpha$ and $\beta$ are coefficients to scale $L_{triplet}$ and $L_{regression}$ to the same unit scale. $f(.)$ is the embedding for the triplet network, $g(.)$ is the output of a regression neural network. $L_{regression}$ can be optimized using Mean Squared Error (MSE) or Mean Absolute Error (MAE). In our experiment, MAE usually leads to better results. The model for combining losses is shown in Fig.~\ref{fig:combining_losses}.

As the model has two loss branches (triplet loss and MOS regression), we can evaluate the ranking in each branch individually. On the triplet loss branch, we choose one image which has the highest MOS rating as reference. The ranking of the test set is then relative to this reference image. On the regression branch, we can predict MOS on the test set and compute the SROCC on the predicted MOS with users' ratings. Both ways give the same ranking accuracy as just using only triplet loss. However, using the regression branch, we do not need to have a reference image to measure the distance to all the test images, but still are able to rank the test images.

\section{Conclusion}

% In this paper, we propose a simple modification from fixed margin triplet loss to an adaptive margin triplet loss. While the original triplet loss is used widely in classification problems such as face recognition, face re-identitication or fine-grained similarity, our proposed loss is well suited for rating datasets in which the ratings are continuous values. In contrast to original triplet loss where we have to sample data carefully, in out method, we can generate triplets using the whole dataset, and the optimization can still converge without frequently running into a model collapsing issue. The adaptive margin only needs to be computed once before the training. It is much less expensive than generating triplets after every epoch as in the fixed margin case. On top of that, our method's performances are also better than the original triplet loss on various rating datasets and network architectures.

In this paper, we present an idea of changing the original triplet loss from fixed margin to adaptive margin. With a simple modification, our proposed algorithm makes triplet losses more usable and extendable from classification datasets to rating datasets. We tested our proposed loss in different datasets, some give big improvements and some have only marginal improvements. Nevertheless, we break the curse of data sampling for triplet training. Our method is also more efficient than the original counterpart by avoiding the re-evaluation on the whole training data after every epoch. The training stability is increased substantially. In fact, we observed no model collapses in our experiments with the proposed model. Therefore, we believe that the idea of adaptive margin triplet loss and its benefit are worth sharing in order to promote the use of triplet training in machine learning.

% Our proposed adaptive margin triplet loss for ranking works well with rating datasets. As we have all the rating values, one natural approach is to train a regression network to predict the ratings. However, the numerical results show that the adaptive margin triplet loss gives better ranking results than regression. By generating triplets for the triplet training, we create more training data than the original data for regression. The relative differences in rating are also preserved in the triplet data.\section{Conclusion}

%\bibliographystyle{splncs}
\bibliographystyle{unsrt}
\bibliography{main}

\begin{thebibliography}{10}

\bibitem{Chopra_Siamese_2005}
Sumit Chopra, Raia Hadsell, and Yann LeCun.
\newblock Learning a similarity metric discriminatively, with application to
  face verification.
\newblock In {\em Proceedings of IEEE Conference on Computer Vision and Pattern
  Recognition}, CVPR '05, pages 539--546, 2005.

\bibitem{Bromley_siamese_1993}
Jane Bromley, Isabelle Guyon, Yann LeCun, Eduard S\"{a}ckinger, and Roopak
  Shah.
\newblock Signature verification using a "siamese" time delay neural network.
\newblock In {\em Proceedings of the 6th International Conference on Neural
  Information Processing Systems}, NIPS'93, pages 737--744, 1993.

\bibitem{Hadsell_siamese_2016}
Raia Hadsell, Sumit Chopra, and Yann LeCun.
\newblock Dimensionality reduction by learning an invariant mapping.
\newblock In {\em Proceedings of IEEE Conference on Computer Vision and Pattern
  Recognition}, CVPR '06, pages 1735--1742, 2006.

\bibitem{Wang_fine_grained_2014}
Jiang Wang, Yang Song, Thomas Leung, Chuck Rosenberg, Jingbin Wang, James
  Philbin, Bo~Chen, and Ying Wu.
\newblock Learning fine-grained image similarity with deep ranking.
\newblock In {\em Proceedings of IEEE Conference on Computer Vision and Pattern
  Recognition}, CVPR '14, pages 1386--1393, 2014.

\bibitem{Sohn_nips_2016}
Kihyuk Sohn.
\newblock Improved deep metric learning with multi-class n-pair loss objective.
\newblock In {\em Proceedings of the 30th International Conference on Neural
  Information Processing Systems}, NIPS'16, pages 1857--1865, 2016.

\bibitem{Schroff_facenet_2015}
Florian Schroff, Dmitry Kalenichenko, and James Philbin.
\newblock Facenet: A unified embedding for face recognition and clustering.
\newblock In {\em Proceedings of IEEE Conference on Computer Vision and Pattern
  Recognition}, CVPR '15, pages 815--823, 2015.

\bibitem{Zhuang_face_2016}
Bohan Zhuang, Guosheng Lin, Chunhua Shen, and Ian Reid.
\newblock Fast training of triplet-based deep binary embedding networks.
\newblock In {\em Proceedings of IEEE Conference on Computer Vision and Pattern
  Recognition}, June 2016.

\bibitem{Ding_face_2015}
Shengyong Ding, Liang Lin, Guangrun Wang, and Hongyang Chao.
\newblock Deep feature learning with relative distance comparison for person
  re-identification.
\newblock {\em Pattern Recognition}, 48(10):2993--3003, Oct 2015.

\bibitem{Hoffer_ranking_2015}
Elad Hoffer and Nir Ailon.
\newblock Deep metric learning using triplet network.
\newblock In {\em Similarity-Based Pattern Recognition}, 2015.

\bibitem{Schwarz_WACV_2018}
Katharina Schwarz, Patrick Wieschollek, and Hendrik P.~A. Lensch.
\newblock Will people like your image? learning the aesthetic space.
\newblock In {\em IEEE Winter Conference on Applications of Computer Vision
  (WACV)}, 2018.

\bibitem{Wu_sampling_2017}
Wu, Smola Chao{-}Yuan, R.~Manmatha, Kr{\"{a}}henb{\"{u}}hl Alexander~J., and
  Philipp.
\newblock Sampling matters in deep embedding learning.
\newblock In {\em IEEE International Conference on Computer Vision, (ICCV)},
  pages 2859--2867, 2017.

\bibitem{AVA}
N.~Murray, L.~Marchesotti, and F.~Perronnin.
\newblock Ava: A large-scale database for aesthetic visual analysis.
\newblock In {\em Proceedings of IEEE Conference on Computer Vision and Pattern
  Recognition}, pages 2408--2415, June 2012.

\bibitem{koniq10k}
Hanhe Lin, Vlad Hosu, and Dietmar Saupe.
\newblock {KonIQ-10K}: Towards an ecologically valid and large-scale {IQA}
  database.
\newblock In {\em arXiv preprint arXiv:1803.08480}, 2018.

\bibitem{ScenicOrNot}
ScenicOrNot.
\newblock Scenic-or-not.
\newblock \url{http://scenicornot.datasciencelab.co.uk}, 2018.

\bibitem{Hermans_defense_2017}
Alexander Hermans*, Lucas Beyer*, and Bastian Leibe.
\newblock {In Defense of the Triplet Loss for Person Re-Identification}.
\newblock {\em arXiv preprint arXiv:1703.07737}, 2017.

\bibitem{Jegou2008_Holiday}
Herve Jegou, Matthijs Douze, and Cordelia Schmid.
\newblock Hamming embedding and weak geometric consistency for large scale
  image search.
\newblock In {\em Proceedings of the 10th European Conference on Computer
  Vision: Part I}, ECCV '08, pages 304--317, 2008.

\bibitem{Pixabay}
Pixabay.com.
\newblock Pixabay dataset.
\newblock \url{https://pixabay.com/}, 2018.
\newblock Accessed: 2018-01-16.

\bibitem{simonyan14_vgg19}
Karen Simonyan and Andrew Zisserman.
\newblock Very deep convolutional networks for large-scale image recognition.
\newblock {\em arXiv preprint arXiv:1409.1556}, 2014.

\bibitem{NIPS2012_AlexNet}
Alex Krizhevsky, Ilya Sutskever, and Geoffrey~E Hinton.
\newblock Imagenet classification with deep convolutional neural networks.
\newblock In {\em Advances in Neural Information Processing Systems 25}, pages
  1097--1105. Advances in neural information processing systems, 2012.

\bibitem{kong2016_AADB}
Shu Kong, Xiaohui Shen, Zhe Lin, Radomir Mech, and Charless Fowlkes.
\newblock Photo aesthetics ranking network with attributes and content
  adaptation.
\newblock In {\em IEEE Conference on European Conference on Computer Vision},
  2016.

\bibitem{Hosu2018-expertise-screening}
Vlad Hosu, Hanhe Lin, and Dietmar Saupe.
\newblock Expertise screening in crowdsourcing image quality.
\newblock In {\em QoMEX 2018: Tenth International Conference on Quality of
  Multimedia Experience}, 2018.

\bibitem{Szegedy2016_Inception}
Christian Szegedy, Vincent Vanhoucke, Sergey Ioffe, Jonathon Shlens, and
  Zbigniew Wojna.
\newblock Rethinking the inception architecture for computer vision.
\newblock {\em IEEE Conference on Computer Vision and Pattern Recognition
  (CVPR)}, pages 2818--2826, 2016.

\end{thebibliography}

\end{document}